\documentclass[letterpaper, 10 pt, conference]{ieeeconf}  

\let\labelindent\relax
\IEEEoverridecommandlockouts                              
\usepackage{cite}
\usepackage{amsmath,amssymb,amsfonts}
\usepackage{graphicx}
\usepackage{textcomp}
\usepackage{xcolor}                                        
\usepackage{algpseudocode}
\usepackage{algorithm}
\usepackage{enumitem}
\title{ Monte Carlo Tree Search Gait Planner for \\Non-Gaited Legged System Control\LARGE \bf}

\author{Lorenzo Amatucci$^{1}$, Joon-Ha Kim$^{2}$, Jemin Hwangbo$^{2}$ and Hae-Won Park$^{2}$  
\thanks{$^{1}$ Lorenzo Amatucci was with the Humanoid Robot Research Center, Korea Advanced Institute of Science and Technology, Daejeon 34141, Korea. He is now with Dynamic Legged Systems Laboratory, Istituto Italiano di Tecnologia (IIT), 16163 Genova, Italy
         \newline
        \indent$^{2}$ Joon-Ha Kim, Jemin Hwangbo and Hae-Won Park are with the Humanoid Robot Research Center, Korea Advanced Institute of Science and Technology, Daejeon 34141, Korea. {\tt\small haewonpark@kaist.ac.kr}}
}

\begin{document}

\maketitle
\thispagestyle{empty}
\pagestyle{empty}

\begin{abstract}
In this work, a non-gaited framework for legged system locomotion is presented. The approach decouples the gait sequence optimization by considering the problem as a decision-making process. The redefined contact sequence problem is solved by utilizing a Monte Carlo Tree Search (MCTS) algorithm that exploits optimization-based simulations to evaluate the best search direction. The proposed scheme has proven to have a good trade-off between exploration and exploitation of the search space compared to the state-of-the-art Mixed-Integer Quadratic Programming (MIQP). The model predictive control (MPC) utilizes the gait generated by the MCTS to optimize the ground reaction forces and future footholds position. The simulation results, performed on a quadruped robot, showed that the proposed framework could generate known periodic gait and adapt the contact sequence to the encountered conditions, including external forces and terrain with unknown and variable properties. When tested on robots with different layouts, the system has also shown its reliability. 

\end{abstract}

\section{INTRODUCTION}

The capabilities of legged animals in dealing with different conditions, such as rough terrains, or effectively react to external disturbances motivated researchers in developing bio-inspired legged systems. However, motion planning and control of this kind of robot represent a difficult challenge since the system's motion results from the interaction between the environment and the feet in contact. Moreover, the body is often underactuated during dynamic gaits, and the contact forces are also constrained due to the physical limitations of the joint actuators and to guarantee non-slipping conditions. 

In most of the previous works as \cite{cheetah3}, \cite{non-linear_so3_MPC} and \cite{frequency-aware_MPC}, a priori knowledge of the motion, such as the future footholds position, the contact foot sequence, or the overall contact timing, are taken as assumptions. Introducing these strategies is necessary to speed up the optimization, especially in real-time systems, but it also limits the variety of the solution. Furthermore, the obtained solution profoundly relies on the handcrafted heuristic required to decouple the gait sequence problem. 

For this reason, various attempts have been made to incorporate contact information in the overall optimization framework. The work in~\cite{Posa} avoided the problem of introducing the contact variables along with the linear complementarity constraint (LCC), but the LCC significantly affects the solver's speed since it does not satisfy the linear independence constraint qualification \cite{NMPC-w-contact}. \cite{TO-w-contact}, and  \cite{NMPC-w-contact} modeled the contact dynamics by inserting spring and damper systems between the ground and the feet. Despite resulting in an explicit contact model with faster solving time, it suffered from the necessity of a trade-off between good gradients in the solver and the physicality of the solution.  In \cite{Winkler}, the LCC is encapsulated in the phased-based parametrization of the ground reaction forces (GRF) and foot position. Although it decreased the solving time significantly, it still was not fast enough for real-time implementation. 

On the other hand, \cite{MIP_Convex} utilized convex mixed-integer formulation to achieve simultaneous contact, gait, and motion planning. They used a binary value to represent the stance-swing phases, the centroidal dynamics model, and fixed phased duration. Consequently, these assumptions limit the solution to only symmetrical gait and have shown problems in more dynamics situations. \cite{gait_selection} dropped the convex representation in favor of a fully nonlinear model of the dynamics integrating the duration of the phases as an optimization variable. This Mixed-Integer Non-Linear Programming (MINLP) formulation is limited to offline use since the average computational time is more than 5 hours. In order to overcome this limitation, the authors trained a neural network to map the resolved MINLP solutions for online gait selection.

Knowledge of the contact state for each foot is fundamental for GRF and footholds optimization since only feet in contact with the ground can contribute to the contact force generated while the feet in the flying phase are free to move to accommodate the base motion. In this work, instead of parameterizing the contact information in continuous variables or as LCC, the problem has been discretized in a decision-making process. For each time step, one in the range of the possible combination of legs in contact is chosen. For a quadruped, this translates to $2^4$ configuration at each time step. The number increases to $16^n$ when the prediction horizon is extended to $n$ time steps. As a result of the combinatorial nature of the problem, the dimension of the search space easily surpasses the computational limit for a real-time brute force search. The absence of a practical heuristic, which can describe the current system status while expanding the tree, made the classic asymmetric growing algorithms, e.g., $A^*$ \cite{chestnutt2005footstep}, challenging. 
 
For this reason, the proposed approach uses the Monte Carlo Tree Search (MCTS) to integrate online contact prediction in an MPC-based control framework. Works like \cite{MCTS_planning} already have shown some of the benefits of MCTS to plan large-scale contact problems involving the reordering of boxes with a robotic arm. 
Furthermore, in artificial intelligence research, the MCTS has shown remarkable performance for solving decision-making problems. One example is the game of Go, in which the high branching factor and the absence of an effective heuristic to evaluate non-terminal state made other algorithms fail~\cite{AlphaGo}. Using MCTS, the Google AlphaGo system reached the super-human performances on the 19x19 board for the first time. 

In our work, integration of MCTS is performed by decoupling the problem from foothold prediction and GRF generation while avoiding the limitation of the handcrafted heuristic besides the MPC cost function. The remainder of the paper is organized as follows. Section \ref{MCTS} explains details of the MCTS algorithm structure used in the proposed algorithm. Section \ref{result} contains the simulation results in various scenarios to show the effect of the proposed algorithm. Section \ref{discussion} discusses the issues about the results. Finally, Section \ref{conclusion} concludes the paper.

\begin{figure}[t!]
    \centering
    \includegraphics[width=0.9\columnwidth]{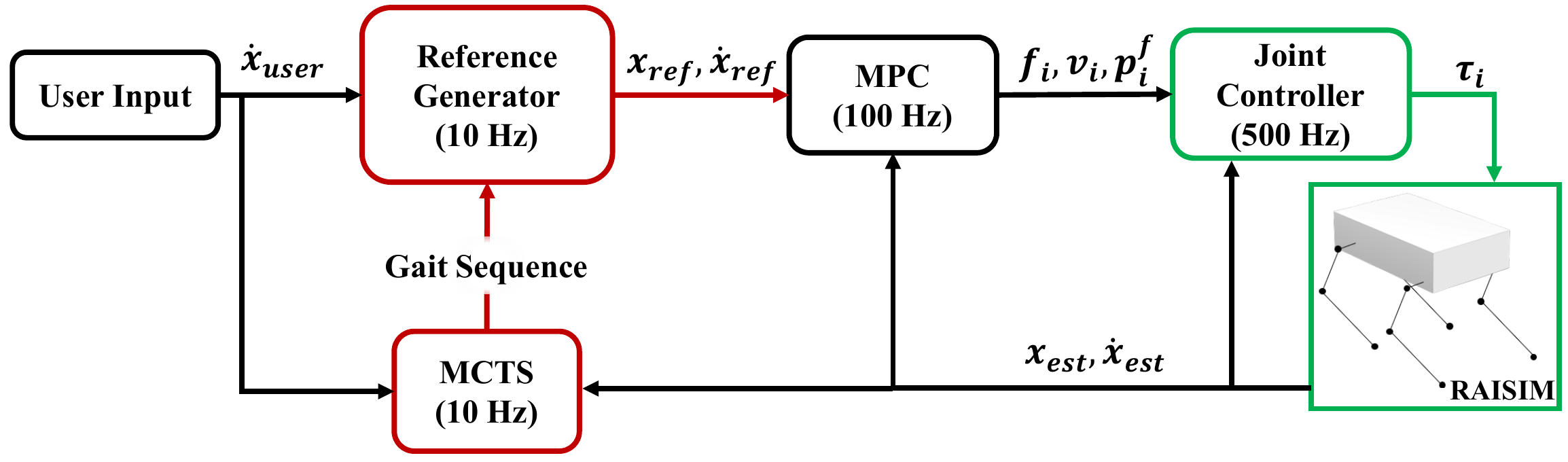}
    \caption[Control framework block diagram]{Diagram of overall control framework. The $\dot{x}_{user}$ is defined as the x, y, yaw directional velocity input from the user. The $x_{est}$ and $\dot{x}_{est}$ is the estimated state values (we used the true values from the simulation in this work).}
    \label{figure: Control Framework}
\end{figure}
\section{MONTE CARLO TREE SEARCH GAIT PLANNER}
\label{MCTS}
The proposed algorithm is presented in Fig. \ref{figure: Control Framework}. The user inputs are $\dot{x}$ and $\dot{y}$, which are respectively the target velocity in the $x$ and $y$ directions, and  $\dot{\psi}$ that is the target yaw rate. The user inputs are then used to generate the reference trajectory for the floating body by integrating the target speed and keeping the other velocities to zero, while the footholds reference are generated using a heuristics as in \cite{foothold_reference}. The MCTS then uses the current contact information and robot states with the reference trajectory to generate the contact sequence used by the MPC.

The MCTS, shown in Algorithm \ref{algo:MCTS}, is an asymmetric growing tree algorithm defined by two different policies, the tree policy and the simulation policy. The tree policy selects the nodes to expand while the simulation policy evaluates the phase sequence defined. The MCTS is employed to optimize the gait sequence utilized in the overall control framework. The algorithm creates a tree search where each node represents one of the possible choices for the contact configuration. Starting from the root node representing the current contact situation, each node deeper in the tree represents the sequence of choices that constitute the gait prediction for the time horizon. At each iteration of the tree growing process, starting from the most promising node, new nodes are expanded considering all the available options (e.g., 16 combination of possible contact configuration for quadrupeds). However, only the options that satisfy the constraint of minimum flying phase duration can be appended to the tree. This minimum flying phase constraint is equal to $0.2s$, which avoids an unwanted bouncing effect of the foot. Each added node continues the simulation by randomly selecting the contact sequence to simulate the considered time horizon. Once the simulations fulfill the considered time horizon length, the simulation policy can be called to evaluate the node. Since the contact sequence is randomly completed, multiple simulations (which is held $n_{sim}$ times as in Table. \ref{tab:sim_parameter}) from each node are run, and the evaluations are averaged. The simulation policy operates by solving a constrained optimization problem to find the optimal system state and control input combination relative to the given contact sequence. The optimization formulation of the simulation policy is the same as the one adopted for the MPC that stabilizes the system motion. Further details are reported in section \ref{simulation policy}. 
\begin{algorithm}[t!]
\caption{MCTS-based gait generator}
\begin{algorithmic}
    \Function{MCTSearch}{$System \_ state ,\: Contact$ }
    \State $ $create  root node$ \:s_0 \: $with$ $
    \State $ $the initial contact state$ \: c_0 $
    \While {$length \: of \:c_{i_{node}} \leq N_{step}$}
    \State $s_{i_{node}} \longleftarrow $TreePolicy$ (s_{i_{node}-1})$
    \State  $ \bar{J}_{i_{node}} \longleftarrow $SimulationPolicy$ (c_{i_{node}})$
    \State  $ $Backpropagation$ (\bar{J}_{i_{node}})$
    \State \Return $ c_{i_{node}} $
\EndWhile
\EndFunction
\end{algorithmic}
\label{algo:MCTS}
\end{algorithm}

\begin{figure}[t!]
\centering
\includegraphics[width=0.9\columnwidth]{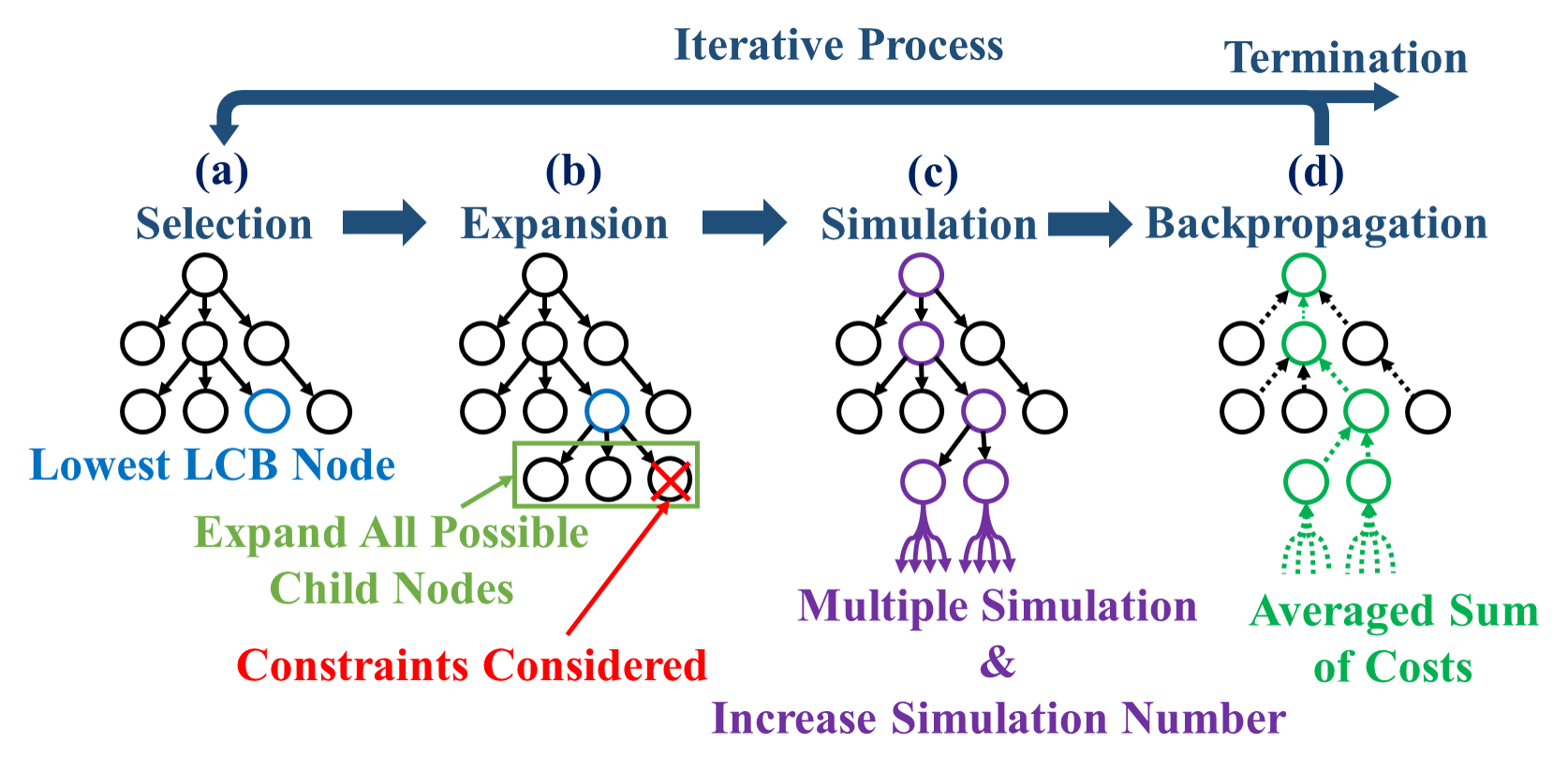}
\caption{Iterative growing process of the MCTS.}
\label{figure:MCTS_scheme}
\end{figure}

After evaluating all the new nodes, the new information acquired through the simulation is backpropagated to the parent node. In this way the evaluation of the parent node is refined. After the tree is updated with the new information, the tree policy traverses the tree until the next node to expand is found. The overall iterative growing process can be synthesized in four steps as shown in Fig. \ref{figure:MCTS_scheme}.

\begin{itemize}[leftmargin=*]
    \item $Selection$: Beginning from the root node, the tree policy is utilized to traverse the tree and select the next node to expand by choosing the node with lowest node value \eqref{eq:LCB}. 
    \item $Expansion$: The selected leaf node is expanded checking each of the feasible phases that can be added to the contact sequence.  
    \item $Simulation$: Multiple simulations are performed, completing the node contact sequence randomly. The node simulation number in \eqref{eq:LCB} is increased for each nodes passed by.
    \item $Backpropagation$: The resulting simulation cost \eqref{eq:node episode cost} from the new simulations of all the child nodes are averaged and assigned to the parent node. The same process is repeated through the branch until the root node.
\end{itemize}
The search terminates when it reaches the maximum computational budget, or the solution converges to a minimum for the considered time horizon as shown in Fig.\ref{figure:MCTS_scheme}.

\subsection{TREE POLICY}
\label{tree policy}
In this subsection, the tree policy of the MCTS is explained in detail. The tree policy affects two of the basic MCTS steps, $Selection$, and $Expansion$. After the simulation of the newly expanded nodes, the acquired information is backpropagated through the tree till reaching the root node. Once the tree is fully updated, the tree policy is responsible for the tree traversal that the following node selection brings. The choice of the next node to expand represents a crucial part of the MCTS algorithm and constitutes the exploitation-exploration dilemma,\cite{Exploitation-exploration}. When deciding the next expansion direction, the algorithm should balance between favoring the most promising node, which means converging faster to a solution or exploring the other possibilities. In order to address these problems, \cite{UCT1} and \cite{UCT2} implemented the ``Upper Confidence Bounds applied to Tree (UCT)". The concept behind this popular method is to compare the upper confidence boundary (UCB) of the nodes evaluation instead of the average itself and pick the one with the higher value. Considering that more confident decisions make narrower confidence intervals, this method increases the possibility that nodes with close evaluation values are expanded while the less promising branches are still discarded for an efficient search. In this work, a variation of the UCB1 algorithm as in \cite{UCB1} is used with the cost in \eqref{eq:node episode cost}. In this algorithm, instead of picking the maximum between the UCB of the node evaluation, the policy selects the minimum between the lower confidence boundary (LCB), which is estimated as follows: 
\begin{equation}
\label{eq:LCB}
LCB_{i_{node}} = \bar{\boldsymbol{J}}_{i_{node}} - c\cdot\sqrt[]{\frac{\log N_{i_{node}}}{n_{i_{node}}}},
\end{equation}
where $\bar{\boldsymbol{J}}_i$ is the average evaluation of the child node, $N_i$ and $n_i$ are respectively the numbers that the parent node and the current $i_{node}^{th}$ node has been simulated, and $c$ is a tunable constant value. This formulation has the advantage of being easy to calculate while coupling the number of times a node is visited and the confidence assigned to the node value estimation. The first term of the expression in \eqref{eq:LCB} is the average evaluation of the node calculated through the simulation of the node itself and its children. It represents the current best estimate of the real node evaluation, indicating the goodness of that expansion direction. The second term represents the confidence in the estimate and is intrinsically linked with the number of times the node has been simulated. The more the considered node is simulated, the smaller the second term of the subtraction results. This implies that the resultant $LCB_{i_{node}}$ is closer to the average evaluation $\bar{\boldsymbol{J}}_{i_{node}}$ for nodes that are visited multiple times, while the less simulated nodes' cost is decreased and giving them an advantage in the $Selection$ process. The $c$ value is empirically defined to balance the two terms and to achieve the desired performance.
\subsection{SIMULATION POLICY}
\label{simulation policy}
\begin{figure}[t!]
\centering
\includegraphics[width=0.55\columnwidth]{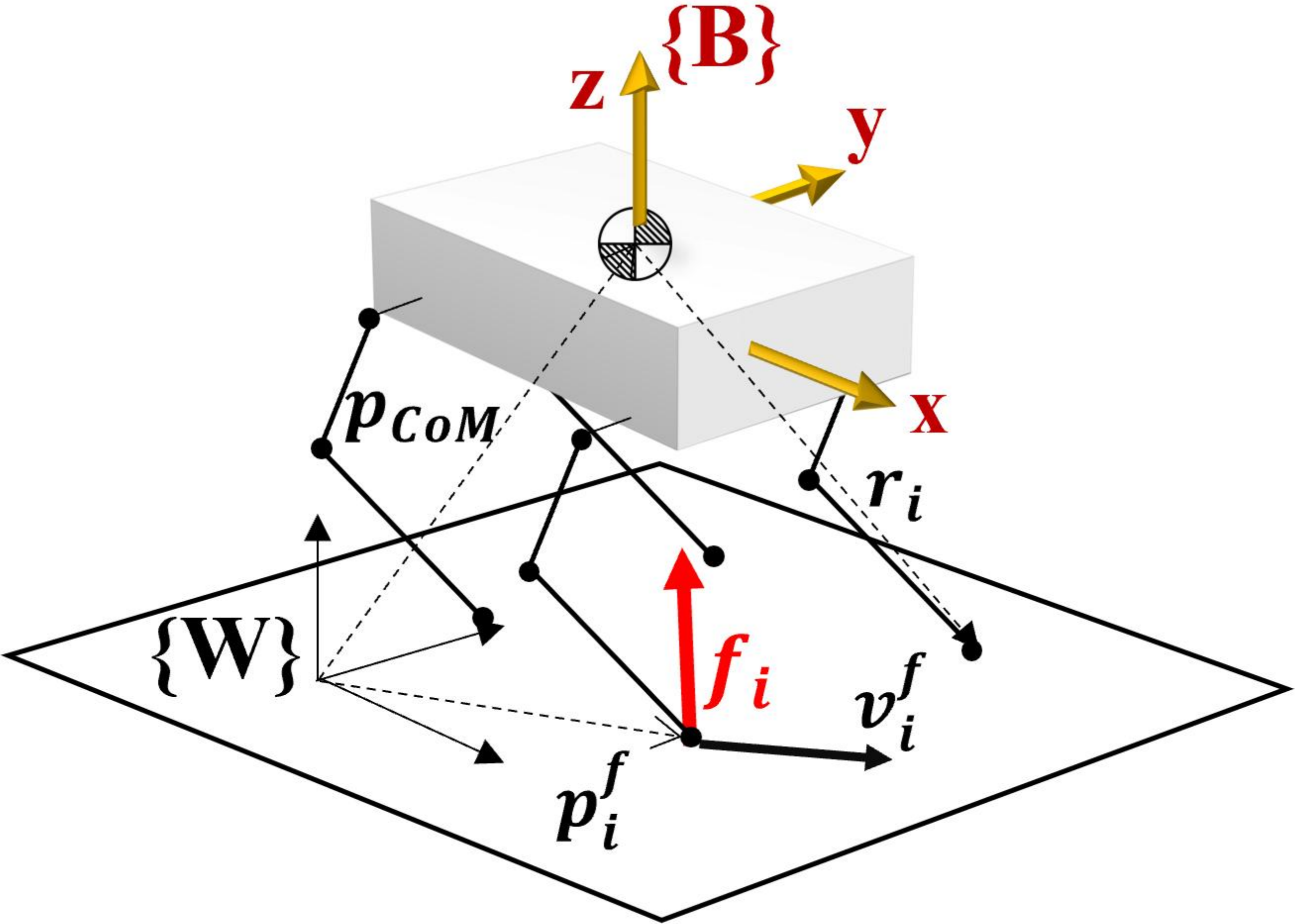}
\caption[3-Dimensional Rigid Body Model]{The adopted robot model and coordinate system. $\boldsymbol{f}_i$ and $\boldsymbol{v}_i^f$ are the control inputs for the $i_{leg}^{th}$ leg, which are respectively the GRF and the horizontal foot speed both expressed in the absolute frame \{W\}.}
\label{figure: 3D rigid body model}
\end{figure}

In this subsection, the simulation policy of the MCTS is explained in detail. The simulation policy uses the same optimization scheme of the MPC-based controllers utilized for the motion stabilization of legged robots, e.g., \cite{RPC}. The optimal control inputs and the future system states are calculated at the same time by the MPC. This framework rapidly simulates the behavior of the robot relative to the chosen gait. The utilized formulation is derivated from \cite{Real-timeMPC}, but the state variable and control input are augmented to include the future foothold optimization. When the assumption of light leg holds, the 3-dimensional single rigid-body model is a reasonable trade-off between the approximation accuracy and the computational efficiency for the dynamics of the robot, as proven by its extensive use in legged system control \cite{cheetah3},\cite{RPC},\cite{frequency-aware_MPC},\cite{linearization}. This assumption eliminates the legs dynamics and their non-linearities from the equation of motion. The rotation matrix representation is chosen to describe the body orientation. While other local formulations like Euler angles offer a more straightforward expression of the equation of motion, they are subjected to singularity configuration that can bring to fatal error in the system control when encountered. A variation-based linearization scheme as in \cite{Real-timeMPC} is used. This scheme assumes that the rotational error expressed in the SO(3) manifold can be formulated considering the variation to the operating point. Utilizing the first-order Taylor expansion of the matrix exponential, rotation error represented in SO(3) is approximated to the rotation variation in $\mathfrak{so}(3)$ and further vectorized for optimization. The correspondent state vector and control vector are defined as: \begin{align}
\boldsymbol{x} &= \left[ \boldsymbol{p}_{CoM},  \dot{\boldsymbol{p}}_{CoM}, \boldsymbol{\xi}, \boldsymbol{\omega}^B, \boldsymbol{p}^f_1, \boldsymbol{p}^f_2, \boldsymbol{p}^f_3, \boldsymbol{p}^f_4\right] \in \mathbb{R}^{24},\nonumber\\
\boldsymbol{u}& = \left[\boldsymbol{f}_1, \boldsymbol{f}_2, \boldsymbol{f}_3, \boldsymbol{f}_4, \boldsymbol{v}^f_1, \boldsymbol{v}^f_2, \boldsymbol{v}^f_3, \boldsymbol{v}^f_4 \right] \in \mathbb{R}^{20},\label{eq: vectorized system state}
\end{align}
where $\boldsymbol{p}_{CoM} \in \mathbb{R}^3$ is the position of the CoM, $\boldsymbol{\xi} \in \mathbb{R}^3$ is the exponential coordinate for the rotation, $\boldsymbol{\omega} \in \mathbb{R}^3$ is the angular velocity vector,   $\boldsymbol{p}_{i_{leg}}^f \in \mathbb{R}^3$ is the position vector of the $i_{leg}^{th}$ foot with $i_{leg} = 1,2,3,4$. The terms $\boldsymbol{f}_{i_{leg}} \in \mathbb{R}^3$ and $ \boldsymbol{v}_i^f \in \mathbb{R}^2$ are respectively the GRF and the horizontal speed of the $i_{leg}^{th}$ foot in a world frame as Fig. \ref{figure: 3D rigid body model}.
In order to complete the formulation of the optimization problem, a cost function needs to be implemented. While the constraints guarantee the feasibility and physicality of the solution, the cost function defines the level of expected performance. Following the scheme utilized in the dynamics linearization, the cost is defined as the weighted sum of the error, on the state and control vector, with respect to the reference. Taking care of the state definition from \eqref{eq: vectorized system state} the cost for the $k^{th}$ prediction is formulated as: 
\begin{equation}
\label{eq:cost function}
\begin{aligned}
    \boldsymbol{\Tilde{J}}_k = &\|\boldsymbol{e}_{\boldsymbol{p}_{CoM},k}\|_{\boldsymbol{Q}_{\boldsymbol{p}_{CoM}}}
	 +\|\boldsymbol{e}_{\dot{\boldsymbol{p}}_{CoM},k}\|_{\boldsymbol{Q}_{\dot{\boldsymbol{p}}_{CoM}}}
	 +\| \boldsymbol{e}_{\boldsymbol{\xi},k} \|_{\boldsymbol{Q}_{\boldsymbol{\xi}}}+\\
	 &+\|\boldsymbol{e}_{\boldsymbol{\omega},k}\|_{\boldsymbol{Q}_{\boldsymbol{\omega}}}
	 +\sum_{i_{leg}=1}^4 \|\boldsymbol{e}_{\boldsymbol{p}_{i_{leg}}^f,k}\|_{\boldsymbol{Q}_{\boldsymbol{p}_{i_{leg}}^f}}+\\
	 &+\sum_{i_{leg}=1}^4 \|\boldsymbol{e}_{f_{i_leg},k}\|_{\boldsymbol{R}_f}+ \sum_{i_{leg}=1}^4 \|\boldsymbol{e}_{\boldsymbol{v}_{i_{leg}}^f,k}\|_{\boldsymbol{R}_{\boldsymbol{v}}},
\end{aligned}
\end{equation}
where $\|\boldsymbol{e}\|_{\boldsymbol{X}} := \boldsymbol{e}^T \boldsymbol{X} \boldsymbol{e}$ and $\boldsymbol{e}_{(.,k)}$ is the error with respect to the reference at the $k^{th}$ prediction. The control input references are defined as $\boldsymbol{f}_{i_{leg},k} =\frac{mg}{N_{leg}}$ ($N_{leg}= 4$ is the number of legs) and $\boldsymbol{v}_{i_{leg},k}^f=[ \dot{x}_{CoM,k},\dot{y}_{CoM,k} ]^{T}$ to minimize the control effort and relative limb velocity. The optimization variables are defined by stacking all the stage variables of each step $k$ in one vector to exploit the state-of-the-art Quadratic Programming solver :
\begin{equation}
    \label{eq: stacked variable}
    \boldsymbol{y} = [\boldsymbol{x}_1,\: \boldsymbol{u}_1,\: ...\quad \boldsymbol{x}_{N_{step}},\: \boldsymbol{u}_{N_{step}}]^T,
\end{equation}
where $N_{step}$ is  the time-horizon length and $\boldsymbol{y} \in \mathbb{R}^{44\cdot N_{step}}$. Based on the new state defined in \eqref{eq: stacked variable}, the linearized system dynamics, the inequality constraints, and the cost function can be manipulated to formulate the overall problem as: 
\begin{equation}
    \label{eq:optimization formulation }
    \begin{aligned}
        \min\limits_{\boldsymbol{y}} \quad &\frac{1}{2}\boldsymbol{y}^T \boldsymbol{W} \boldsymbol{y} + \boldsymbol{y}^T \boldsymbol{h}.\\
         s.t. \quad & \boldsymbol{A}_{eq}\boldsymbol{y}=\boldsymbol{b}_{eq},\\
         &\boldsymbol{A}_{ineq}\boldsymbol{y}\leq \boldsymbol{b}_{ineq},
    \end{aligned}
\end{equation}
where $\boldsymbol{W} \in \mathbb{R}^{(44 \cdot N_{step}) \times (44 \cdot N_{step})}, \boldsymbol{h} \in \mathbb{R}^{44 \cdot N_{step}}$ come from the cost function, $\boldsymbol{A}_{eq} \in \mathbb{R}^{(n_{eq} \cdot N_{step})\times(44 \cdot N_{step})}$, $\boldsymbol{b}_{eq} \in \mathbb{R}^{n_{eq} \cdot N_{step}}$, are defined by the linearized system dynamics and $\boldsymbol{A}_{ineq} \in \mathbb{R}^{(n_{ineq} \cdot N_{step}) \times (44 \cdot N_{step})}$ and $\boldsymbol{b}_{ineq} \in \mathbb{R}^{n_{ineq} \cdot N_{step}}$ are the linearized friction cone constraint and box constraint imposed to guarantee the feasibility of the solution and to ensure the non-slipping condition ($n_{eq}$ and $n_{ineq}$ indicates the number of equatity and inequality contraints per each step, respectively) further details on the matrix construction are in \cite{Real-timeMPC}. $qpSWIFT$ \cite{qpSWIFT} is chosen among other solvers to solve the optimization due to its solution speed with the ability to exploit the sparse structure of the problem. By utilizing $qpSWIFT$ in the simulation process of the MCTS gait generator, the system can perform simulations up to 250 Hz without including any parallelization techniques.


Now, with the solved optimal control inputs and states, the performance of the gait is evaluated with the cost function $\boldsymbol{\Tilde{J}}_k$, defined in equation \eqref{eq:cost function}, with the addition of a new term related to the contact sequence. The additional cost gives more freedom to tune the solution to maximize the number of legs in contact at each step. The final node episode cost $\boldsymbol{J}$ is then defined as follows:
\begin{equation}
\label{eq:node episode cost}
 \boldsymbol{J} = \sum_{k = 1}^{N_{step}} \{  \boldsymbol{\Tilde{J}}_k + R_c\cdot(N_{leg}-\sum_{i_{leg} = 1}^{N_{leg}}c_{i_{leg},k})\},
\end{equation}
where $R_c$ is the weight relative to the contact cost, and $c_{i_{leg},k}$ is the binary variable representing the contact state for the $i^{th}$ leg in $k^{th}$ step. The new term in equation \eqref{eq:node episode cost} minimize the leg's air-time, increasing the stability of the solution at lower target speeds. 

\section{SIMULATION RESULT}
\label{result}
\begin{table}[t!]
\caption{Robot and framework parameter}
\label{tab:sim_parameter}
\begin{center}
\begin{tabular}{c c c c}
\hline
     Parameter & Value  & Parameter & Value\\
     \hline
     $m$ & 19 kg   & $\mu$  & 0.7\\
     $\boldsymbol{I}$ & $diag$(1e-2[9 60 67])$Kgm^2$ & c & 1.5 \\
     $\boldsymbol{Q}_{\boldsymbol{p}_{CoM}}$& $diag$(1e3[1 1 30]) & $n_{sim}$ & 9 \\
     $\boldsymbol{Q}_{\boldsymbol{\dot{p}}_{CoM}}$ & $diag$(1e1[10 1 1])& $dt_{tree}$ & 0.1 s\\
     $\boldsymbol{Q}_{\boldsymbol{\xi}}$ & $diag$(1e3[2 2 3])& $ dt_{MPC}$ & 0.02 s\\
     $\boldsymbol{Q}_{\boldsymbol{\omega}}$ & $diag$(1e1[1 1 1]) & $T_{MPC}$ & 0.4 s \\
     $\boldsymbol{Q}_{\boldsymbol{p}_{i_{leg}}}$ & $diag$(1e3[1 1 0]) & $Body \: Length$ & 0.6 m \\
     $\boldsymbol{R}_f$ & $diag$(1e-3[1 1 1]) & $Body \: Width$ & 0.2 m \\
     $\boldsymbol{R}_{\boldsymbol{v}}$ & $diag$(1e1[1 1]) & $T_{tree}$ & 0.6 s\\
     \hline
\end{tabular}
\end{center}
\end{table}
In this section, the proposed algorithm is verified on various simulation environments with a quadrupedal robot using the Raisim \cite{raisim} simulator, where the parameters of the robot and the framework are summarized in Table \ref{tab:sim_parameter}. The joint torques values to actuate the robot are calculated as follows by using GRFs, feet positions and velocities, which are generated by the MPC:
\begin{align}
\pmb{ \tau }_{i_{leg}} = \boldsymbol{J}_{i_{leg}}^T[&\boldsymbol{c}_{i_{leg},k}\boldsymbol{f}_{i_{leg}}+(1-\boldsymbol{c}_{i_{leg},k})\boldsymbol{k}_p(\boldsymbol{p}^f_{i_{leg},est}-\boldsymbol{p}^f_{i_{leg}}) \nonumber\\
&+(1-\boldsymbol{c}_{i_{leg},k})\boldsymbol{k}_d(\dot{\boldsymbol{p}}^f_{i_{leg},est}-\dot{\boldsymbol{p}}^f_{i_{leg}})],
\end{align}
where $\boldsymbol{J}_{i_{leg}}$ indicates the leg kinematic Jacobian, $\boldsymbol{k}_p$ and $\boldsymbol{k}_d$ indicates the tunable Proportional and Differential (PD) control gain, $\boldsymbol{p}^f_{i_{leg},est}$ and $\dot{\boldsymbol{p}}^f_{i_{leg},est}$ is the estimated foot position and velocity and  $\boldsymbol{\dot{p}}^f_{i_{leg}} = [{\boldsymbol{v}^f_{i_{leg}}}^{T},\dot{z}_{i_{leg}}]^{T}$ where $\dot{z}_{i_{leg}}$ is the speed in the $z$ direction as a $5^{th}$ order polynomial.

\subsection{MCTS COST ANALYSIS}
\begin{figure}[t!]
   \centering
    \includegraphics[width=0.85\columnwidth]{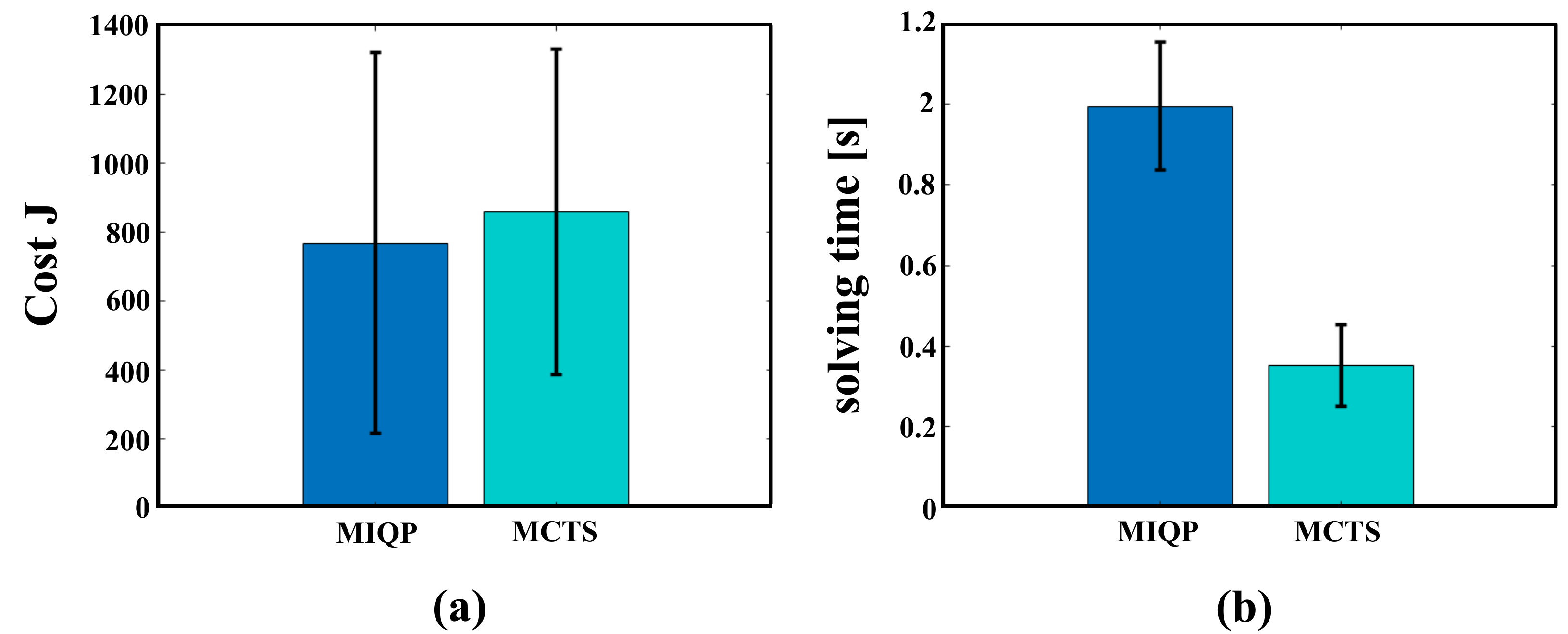}
    \caption[MIQP vs MCTS]{The average cost of the MIQP and the MCTS on graph (a) and the relative solving time for the two algorithms on graph (b).}
    \label{fig:MIQPvsMCTS}
\end{figure}

Due to the intrinsic implementation of the contact as a binary variable, it is natural to compare the proposed control framework with a Mixed Integer Quadratic Programming (MIQP) formulation. MIQP formulations have already been successfully used in various works for the contact planning of biped robots as in \cite{MIQP_planning_biped_1}, and \cite{MIQP_planning_biped_2}, as well as for quadrupeds in \cite{MIP_Convex}. To obtain a fair comparison, the same constraints, dynamics linearization, and cost function presented in section \ref{simulation policy} are applied. Also, the limitation of $0.2 s$ minimum swing time is introduced as in the MCTS. A state-of-the-art optimization solver GUROBI, \cite{gurobi}, is utilized to solve the proposed optimization problem. Fig. \ref{fig:MIQPvsMCTS}. shows the average cost of the MCTS and MIQP with the respective standard deviations. The values are obtained by averaging the cost calculated by the two algorithms on the same robot conditions with the same reference and time horizon. Multiple simulations are run with different target speed from $0 m/s$ to $2.5 m/s$ with random external forces applied with an average magnitude of $30N$. As reported in Fig. \ref{fig:MIQPvsMCTS}., the average cost of the MCTS is only $10\%$ bigger than the one of the MIQP while having 2.72 times faster solution. The comparison shows the consistency of the MCTS gait sequence generator in terms of cost minimization and the computational advantages of this formulation with respect to the state-of-the-art solver.
\begin{figure}[b!]
    \centering
    \includegraphics[width=0.65\columnwidth]{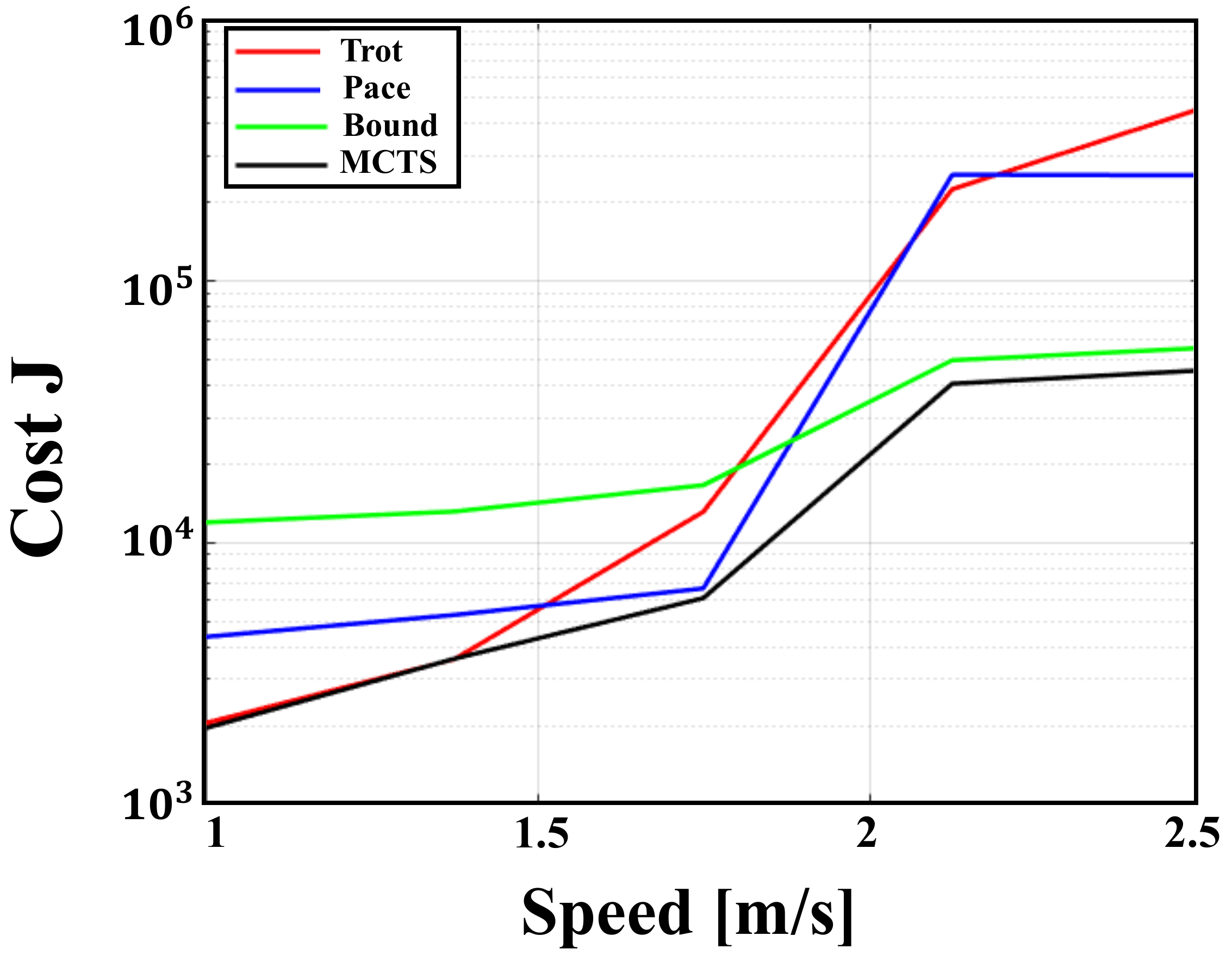}
    \caption[Gait cost over target speed]{Cost comparison of MCTS with predefined gaits over target speed variation. The x-axis is the target speed while the y axis, in logarithmic scale, is the average running cost of the MPC evaluated at steady state.}
    \label{figure:speed_comparison}
\end{figure}
\subsection{GAIT GENERATION ANALYSIS}
The advantages of having the possibility of gait adaptation can be seen by analyzing the MPC cost at different speeds. Fig. \ref{figure:speed_comparison}. shows the average cost calculated by the MPC during the simulations. The gait generated from MCTS and predefined gaits is compared with different target speeds that varies from $1 m/s$ to $2.5 m/s$. We could find out that MCTS outperforms predefined gaits. The costs are recorded for $3s$ to evaluate the reached limit cycle. The comparison is performed with three different predefined gaits: Trot, Pace, and Bound. The MPC cost is chosen as a metric for the evaluation since it is tuned to represent an optimal overall performance for the robot at the given condition and target input.
\begin{figure}[t!]
    \centering
    \includegraphics[width=0.8\columnwidth]{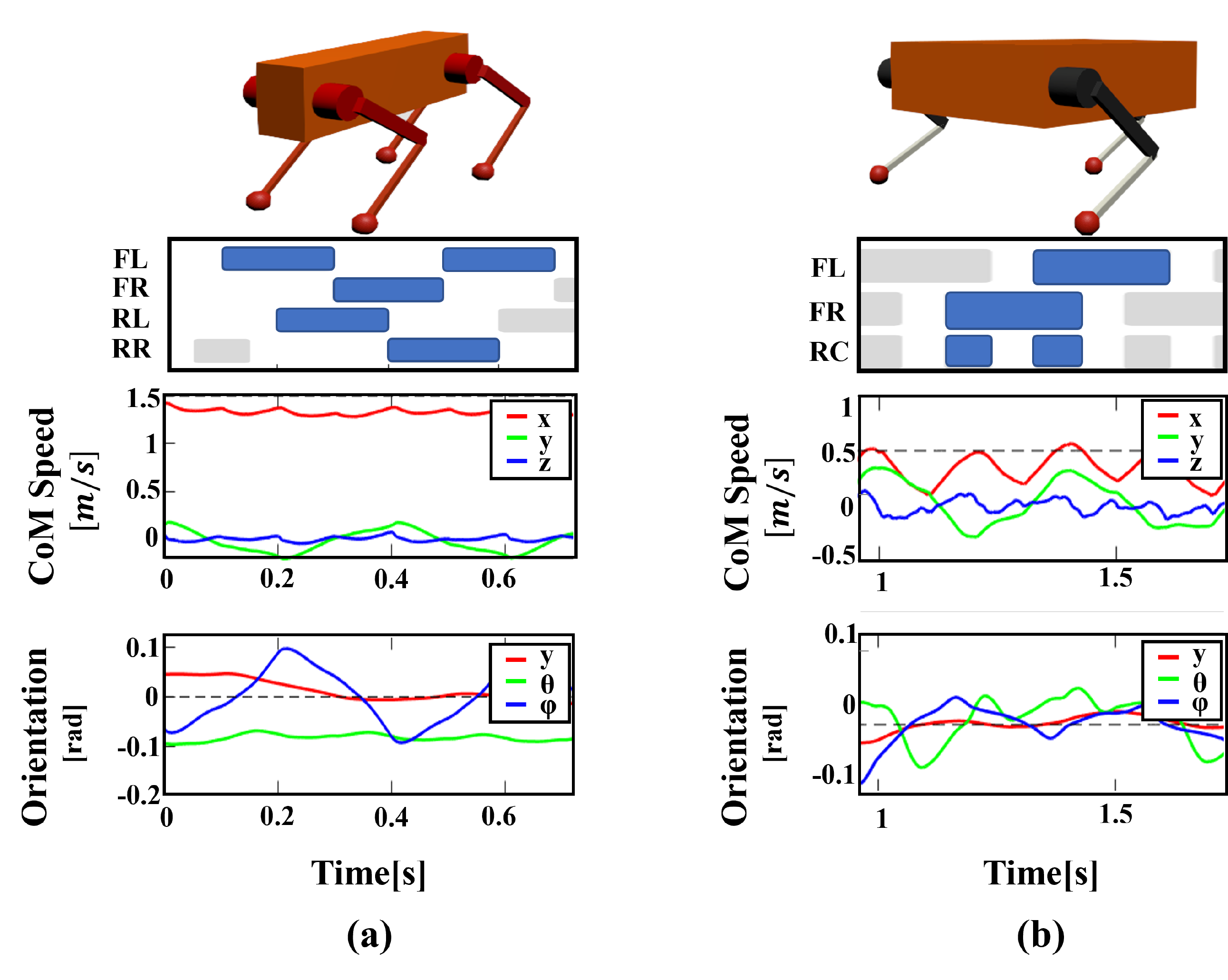}
    \caption[MCTS Gait]{Example gait for quadruped robot generated at a target speed of $1.5m/s$ on graph (a). The example gait generated for the three leg robot at a target speed of $0.5 m/s$ on graph (b). RC stands for the Rear Center leg.}
    \label{figure: MCTS_gait}
\end{figure}
As written in Table \ref{tab:sim_parameter}, the MCTS works with a time horizon of $0.6 s$ and a frequency update of $10 Hz$. Interestingly, for a target speed of $1.5 m/s$, if the optimal contact sequence is recalculated at each framework's cycle by MCTS, the controller reaches a limit cycle as shown in Fig. \ref{figure: MCTS_gait}. One thing to consider is that the limit cycle period length and the MCTS time horizon length are not linked and the algorithm has proven to discover periodic cycles with shorter or longer periods than the considered time horizon.
\begin{figure}[t!]
    \centering
    \includegraphics[width=1\columnwidth]{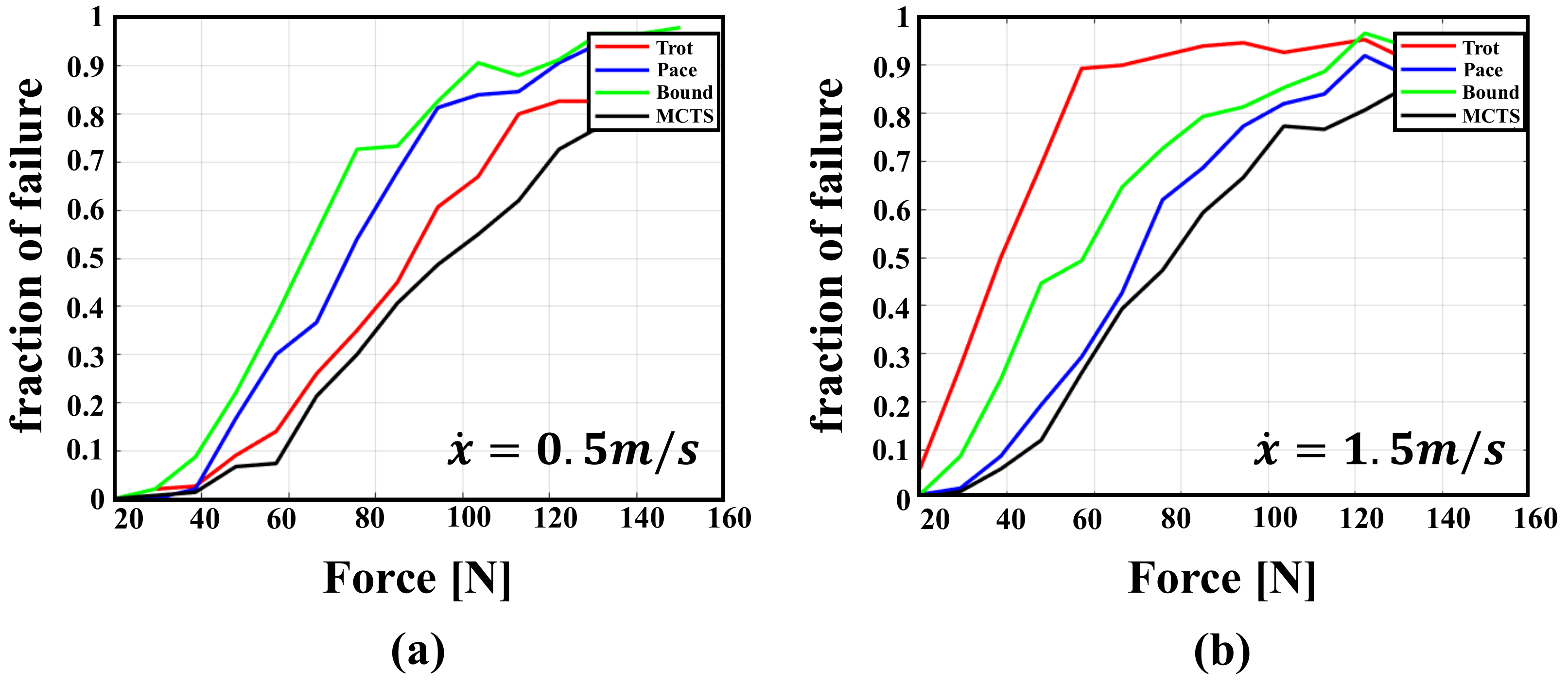}
    \caption[System Robustness ]{Fraction of failed simulation at a defined force magnitude with the system target speed at $0.5m/s$ on (a) and $1.5m/s$ on (b).}
    \label{figure:Robusteness}
\end{figure}

\subsection{Robustness}
One of the advantages of the non-gaited locomotion of the system is the robustness of the robot against external disturbances. This is tested in simulation by generating a pattern of random external forces applied to the robot body. The force is applied multiple times in randomized directions, with an average of $90 \deg $ difference from the direction of the motion. The test is failed if the robot touches the ground with a part that is not the foot. The simulations are performed with a target speed of $0.5 m/s$ and $1.5 m/s$. The system is compared with the same fixed gait utilized in the previous section to analyze the gait difference at various speeds. Fig. \ref{figure:Robusteness} reports the frequency of failure with respect to the force magnitude on a sample of 250 simulations. The MCTS always achieves better performances than the fixed gait with a decrease in the fraction of failure up to $13\%$.
\begin{figure}[t!]
    \centering
    \includegraphics[width=0.7\columnwidth]{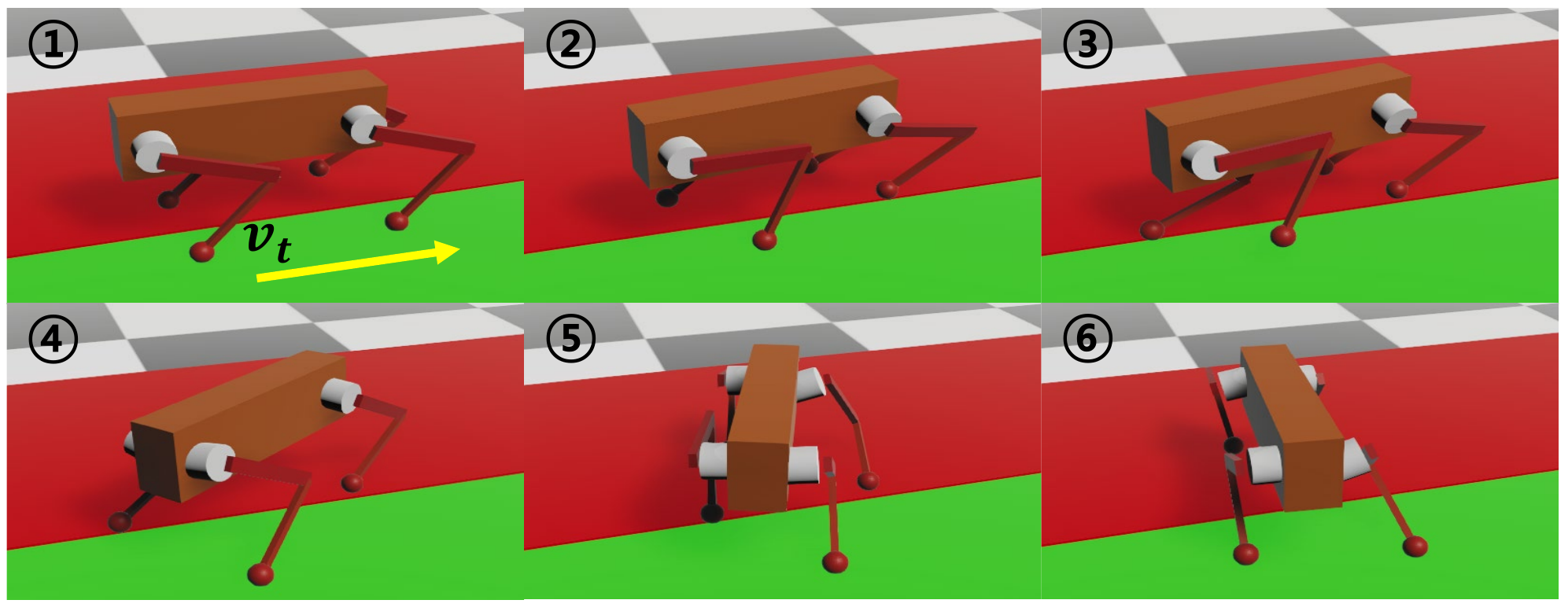}
    \caption[Spinning on the two speed treadmill]{Simulation of the quadruped robot spinning in $\dot{\psi}$ = $1 rad/s$, while the treadmill is moving under its feet with $v_t=0.2 m/s$.}
    \label{figure: spin_treadmill}
\end{figure}
\begin{figure}[t!]
    \centering
    \includegraphics[width=0.7\columnwidth]{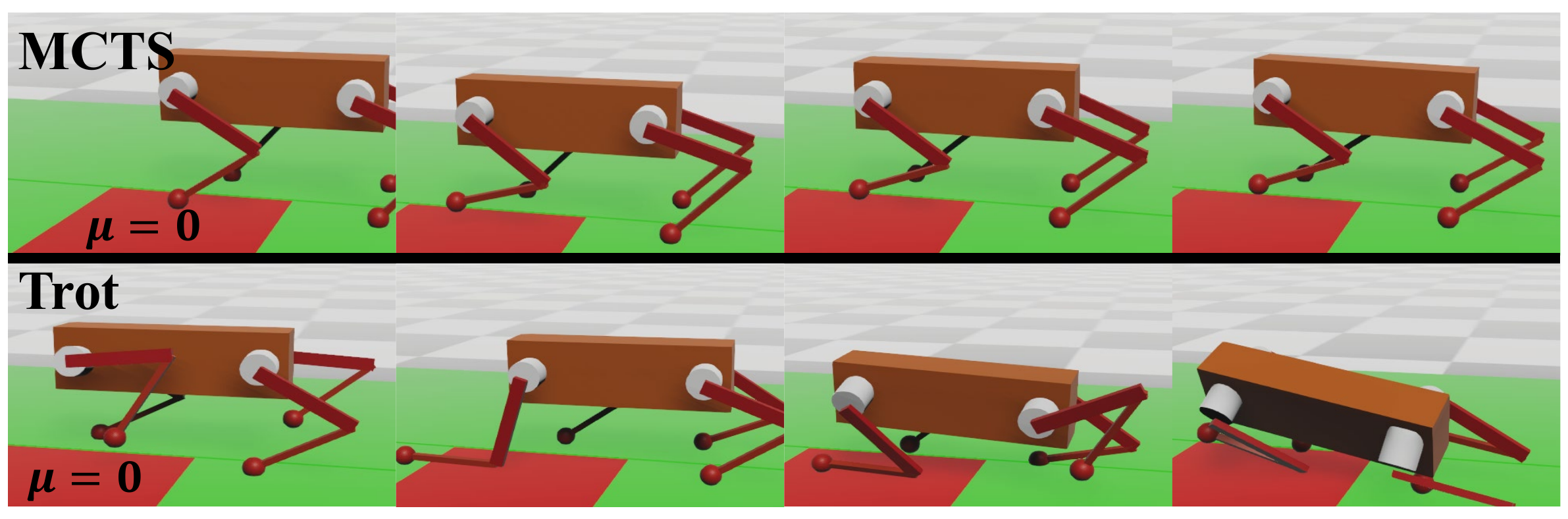}
    \caption[Standing on slippery terrain]{Example of standing on a ground with one leg over a surface with no friction (red region). Top row is the simulation performed with the MCTS, and the bottom utilizing the trot gait. The time moves forward form left to right.}
    \label{figure:screen_stance_slip}
\end{figure}
\begin{figure}[t!]
    \centering
    \includegraphics[width=0.9\columnwidth]{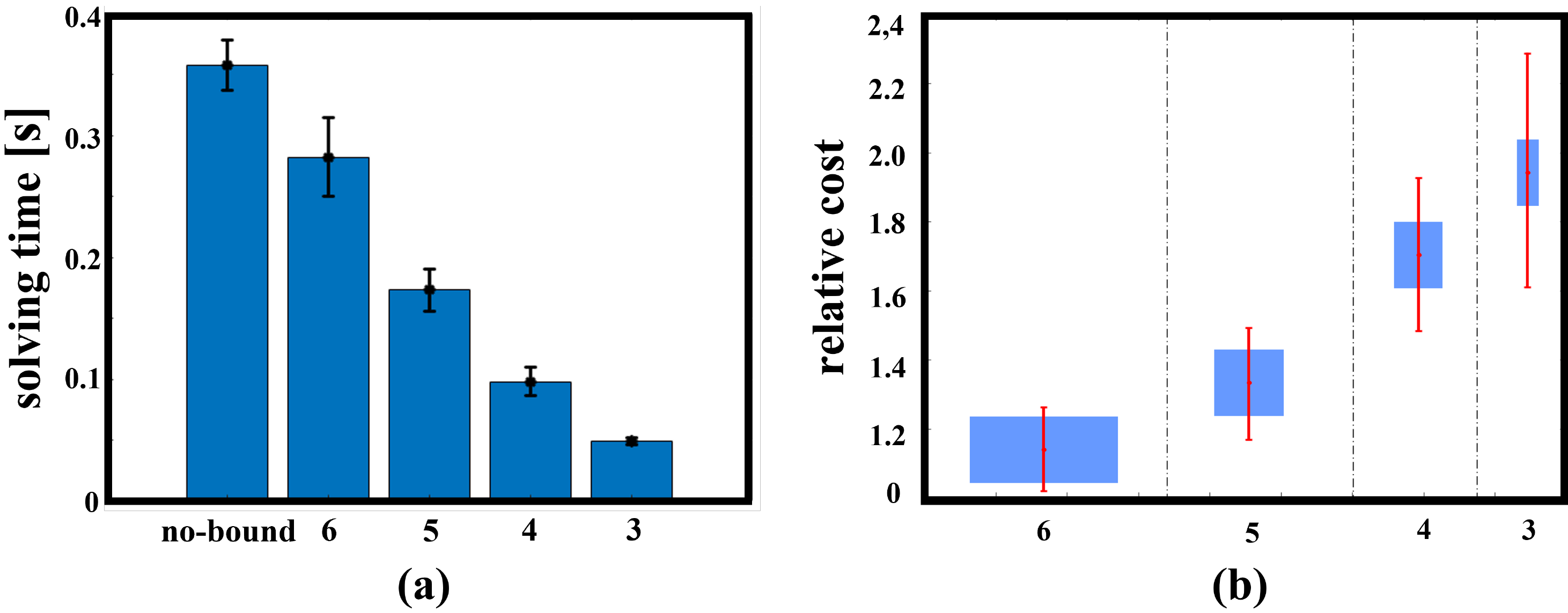}
    \caption[Cost-time comparison with limited tree expansion]{Comparison between 4 different runs of the MCTS algorithm limited to 3, 4, 5, and 6 tree expansion iteration. The performances are compared for the solving time on (a) and for the relative cost difference with the non-bounded algorithm on (b).}
    \label{figure: error_eval}
\end{figure}
Another advantage of the non-gaited locomotion of the system is the robustness against an unknown environment. In a model-based framework, various assumptions are taken to model the contact with the ground. When these assumptions are not satisfied, the system can irremediably fail. The proposed framework based on the MPC controller and enhanced with the MCTS gait generator can overcome these problems without including any terrain information inside the model. Adequate changes in the contact sequence can mitigate the effect of unexpected terrain conditions and optimize the efficiency of the consequent reaction. The robot is tested on two treadmills, one per side, to set different speeds for each part of the robot. As shown in Fig. \ref{figure: spin_treadmill}. a target yaw rate, $\dot{\psi}$, is given as input while the green treadmill is moving with a speed of $v_t = 0.2m/s$ and the red one is kept still. The screenshot in Fig. \ref{figure:screen_stance_slip}. also shows the system's performance when one of the legs is on a slippery terrain. The generated gait is compared with the performance achieved utilizing a fixed gait controller that cannot stabilize the system and maintain the base position. The MCTS does not have any additional information about the environment, but the leg position feedback is enough for the algorithm to generate a gait that can stabilize the system.
\subsection{MODIFIED LEGGED ROBOT}
For a quadruped robot, the nature gives us a hint for designing the contact sequences. However, it is not easy to design a predefined gait sequence for a robot with an unnatural number and configuration of legs. Tuning the stance and swing time for this system is a tedious job and requires intuition and much testing to find a gait that can perform the desired task. Thankfully, the proposed framework's only necessary change is to specify the number of legs of the desired system and the hip position to create a compatible contact sequence. For example, a robot with two legs of the front and only one on the rear is tested. As the right graph of Fig. \ref{figure: MCTS_gait}. shows, the framework has no problem in finding a successful gait sequence for this leg configuration. The example highlights a periodic gait discovered by the system when the user imposes a target speed of $0.5m/s$. It should be noticed that the rear leg makes twice the contact with respect to the front legs due to the nature of leg configuration.
\section{DISCUSSION}
\label{discussion}
The presented work is implemented in MATLAB, and the shown simulations are performed on a six-core mobile processor, the Intel(R) i7-9750H. Due to current implementation limitations, the MCTS is not able to run in real-time. In the presented results, the average time for the solution to converge is $0.37s$ while the MCTS performed, on average, 92 simulations to predict six time-steps. Considering the property of the MCTS, the algorithm can be interrupted before convergence while still getting feasible results for an increase in speed up to $97 \%$.  Further details can be seen in Fig. \ref{figure: error_eval}. where the costs and solving times at different expansion cycles are reported. Furthermore, various work as \cite{parallel_MCTS_1} and \cite{parallel_MCTS_2} showed the benefit of introducing parallelization techniques in the MCTS framework to speed up the computation. The algorithm can also speed up by substituting the optimization-based simulation with a neural network imitating the simulation in~\cite{mpc_net}. 
\section{CONCLUSION}
\label{conclusion}
In this paper, a novel framework for non-gaited legged locomotion control has been presented. The proposed method exploits the formulation of the contact sequence generation as a decision-making problem. The contact sequence optimization is solved with a novel MCTS-based approach. The gait generation is decoupled from the GRF and foothold optimization carried by the MPC-based controller. The modified MCTS uses the prediction capabilities of the MPC to explore the vast search space of possible phases combination to give output as the contact sequence over a fixed time horizon. The shown comparison in simulation environments highlights the potential of this framework against the state-of-the-art MIQP solvers. The presented simulation result proves that the proposed method can automatically discover periodic gaits and adapt to external disturbances and unknown terrain morphology and characteristics, increasing the system's robustness. Finally, the framework is easily adaptable to various robot layouts.




\bibliographystyle{ieeetr}
\bibliography{library.bib}
\end{document}